%% file: main.tex
\def\year{2021}\relax
\documentclass[letterpaper]{article} 
\usepackage{aaai21}  
\usepackage{times}  
\usepackage{helvet} 
\usepackage{courier}  
\usepackage[hyphens]{url}  
\usepackage{graphicx} 
\usepackage{graphicx}  
\usepackage{natbib}  
\usepackage{caption} 

\usepackage{amsmath}
\usepackage{xcolor}
\newcommand{\etal}{\textit{et al}.}
\newcommand{\etc}{\textit{etc}}
\newcommand{\ie}{\textit{i}.\textit{e}.}

\usepackage[switch]{lineno} 
\newcommand{\jx}[1]{#1}
\newcommand{\kw}[1]{#1}

\newcommand{\nr}[1]{#1}
\newcommand{\jm}[1]{#1}

\usepackage{mathtools}

\DeclarePairedDelimiterX{\infdivx}[2]{(}{)}{%
  #1\;\delimsize\|\;#2%
}
\newcommand{\infdiv}{D_{KL}\infdivx}

\title{\jx{Encoding Syntactic Knowledge in Transformer Encoder \\ for Intent Detection and Slot Filling} }
 \author{
    Jixuan Wang\textsuperscript{\rm 1}\thanks{Work done during author's internship at Amazon Alexa.},
    Kai Wei\textsuperscript{\rm 2}\thanks{Corresponding author.},
    Martin Radfar\textsuperscript{\rm 2}, 
    Weiwei Zhang\textsuperscript{\rm 2},
    Clement Chung\textsuperscript{\rm 2}
    \\
}
\affiliations {
    \textsuperscript{\rm 1} University of Toronto, Vector Institute, 
    \textsuperscript{\rm 2} Amazon Alexa \\
    jixuan@cs.toronto.edu, \{kaiwe, radfarmr, wwzhang, chungcle\}@amazon.com
}
\begin{document}

\maketitle

\begin{abstract}
\input{abstract.tex}
\end{abstract}

\section{Introduction}
\input{introduction.tex}

\section{Problem Definition}
\input{problem_definition.tex}

\section{Proposed Model}
\input{proposed_model.tex}

\section{Experiments}
\input{experiment.tex}

\section{Results}
\input{results.tex}

\section{Related Work}
\input{related_work.tex}
\section{Conclusion}
\input{conclusion.tex}

\section{Acknowledgement}
We would like to thank Siegfried Kunzmann, Nathan Susanj, Ross McGowan, and anonymous reviewers for their insightful feedback that greatly improved our paper. 
\bibliography{ref}

\section{Appendix A}
Below lists examples of the intent detection errors made by the model without syntactic information that are related to one specific grammar pattern between prepositions and nouns. 
\begin{itemize}
    \item cleveland to kansas city arrive monday before 3 pm 
    \item kansas city to atlanta monday morning flights
    \item new york city to las vegas and memphis to las vegas on Sunday
\end{itemize}

Below lists examples of the slot filling errors made by the model without syntactic information that contain POS confusion. 
\begin{itemize}
    \item cleveland to kansas city arrive monday before 3 pm 
    \item new york city to las vegas and memphis to las vegas on Sunday
    \item baltimore to kansas city economy
\end{itemize}

The Transformer encoder-based model without syntactic information made mistakes on all these utterances. The model trained with POS tagging and the model trained with both POS tagging and dependency prediction fail on the last utterance in the list below. The model trained with dependency prediction does not make any mistakes on all these utterances. We underline the words that are assigned to wrong slots by the model without syntactic information. 
\begin{itemize}
    \item book a	reservation for velma	 an a  and rebecca for an american pizzeria \underline{at} (correct: $B-TimeRange$; prediction: $B-RestaurantName$) $5$ Am in MA
    \item Where is Belgium \underline{located} (correct: $Other$; prediction: $B-PatialRelation$)
    \item \underline{May}(correct: $Other$; prediction: $B-TimeRange$)  I have the movie schedules for Speakeasy Theaters
\end{itemize}

\end{document}

%% file: abstract.tex
We propose a novel Transformer encoder-based architecture with syntactical knowledge encoded for intent detection and slot filling. Specifically, we encode syntactic knowledge into the Transformer encoder by jointly training it to predict syntactic parse ancestors and part-of-speech of each token via multi-task learning. Our model is based on self-attention and feed-forward layers and does not require external syntactic information to be available at inference time.
Experiments show that on two benchmark datasets, our models with only two Transformer encoder layers achieve state-of-the-art results. Compared to the previously best performed model without pre-training, our models achieve absolute F1 score and accuracy improvement of $1.59\%$ and $0.85\%$ for slot filling and intent detection on the SNIPS dataset, respectively. Our models also achieve absolute F1 score and accuracy improvement of $0.1\%$ and $0.34\%$ for slot filling and intent detection on the ATIS dataset, respectively, over the previously best performed model. Furthermore, the visualization of the self-attention weights illustrates the benefits of incorporating syntactic information during training.

%% file: introduction.tex

\begin{figure}[t]
    \centering
    \includegraphics[width=0.48\textwidth]{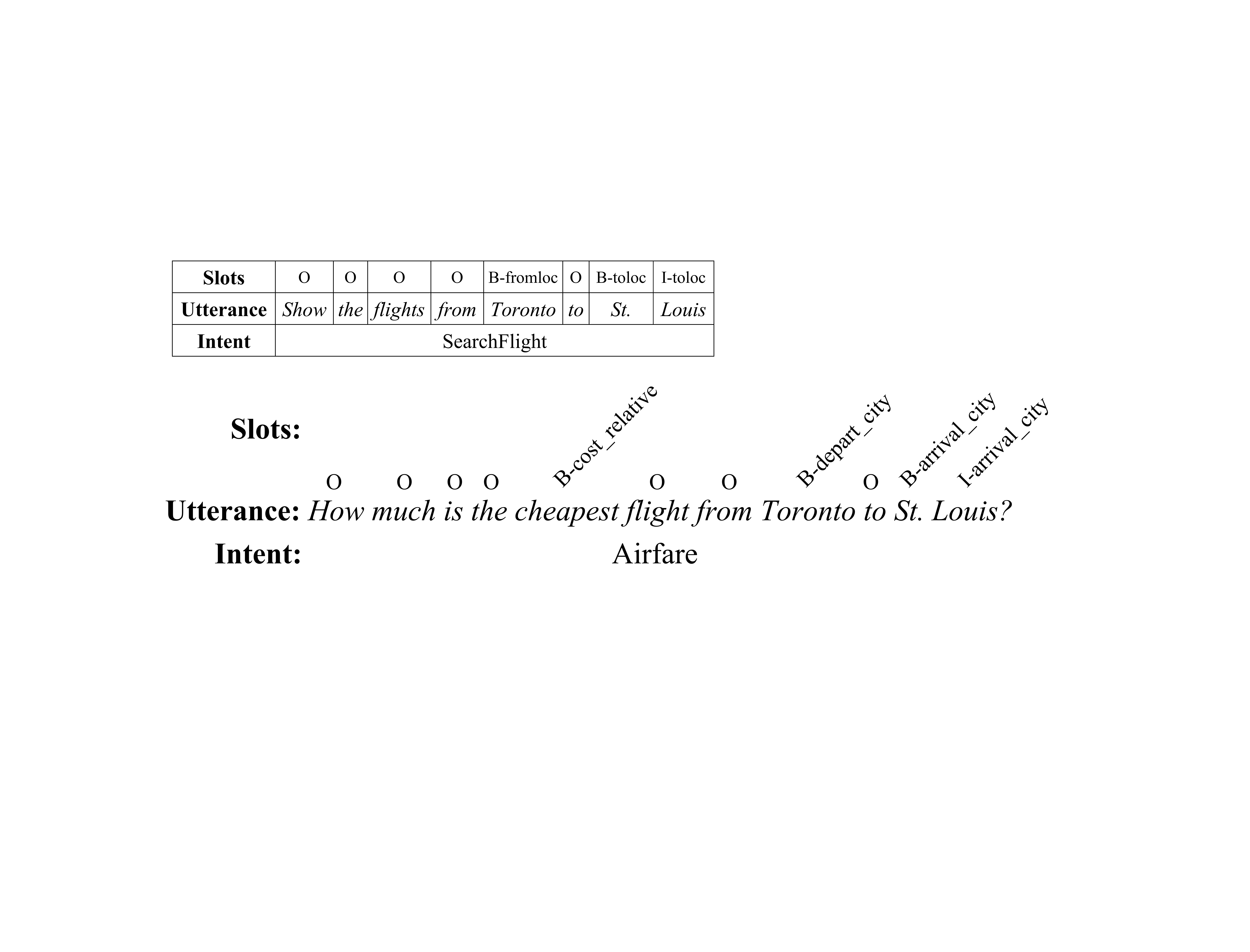} 
    \caption{An example of NLU tasks.}
    \label{fig:example}
\end{figure}

Recent years have seen great success in applying deep learning approaches to enhance the capabilities of virtual assistants (VAs) such as Amazon Alexa, Google Home and Apple Siri. One of the challenges for building these systems is \nr{mapping the meaning of users’ utterances, which are expressed in natural language, to machine comprehensible language~\cite{allen1995}}. An example is illustrated in Figure~\ref{fig:example}. In this utterance ``\emph{Show the cheapest flight from Toronto to St. Louis}'', the machine needs to map this utterance to an intent \emph{Airfare} (intent detection) and to slots such as \emph{Toronto: FromLocation} (slot filling). In this work, we focus on intent detection and slot filling and refer to these as Natural Language Understanding (NLU) tasks.



Previous works show that a simple deep neural architecture delivers better performance on 
NLU tasks when compared to traditional models such as 
Conditional Random Fields~\cite{collobert2011natural}.
Since then, deep neural architectures, predominantly recurrent neural networks, have become an indispensable part of building NLU systems~\cite{zhang2016joint, goo2018slot, e2019a}.
Transformer-based architectures,
as introduced more recently 
by~\cite{vaswani2017attention}, have 
shown significant improvement over previous 
works on NLU tasks~\cite{ chen2019bert, qin2019a}.
\kw{Recent studies 
show
that although the Transformer model
can learn
syntactic 
\jx{knowledge}
purely by seeing examples, explicitly feeding this 
\jx{knowledge}
to such models can significantly enhance their performance on tasks such as neural machine translation~\cite{sundararaman2019syntax}
and semantic role labeling~\cite{strubell2018linguistically}. }
\kw{While incorporating syntactic knowledge has been shown to improve performance for NLU tasks~\cite{tur2011sentence, chen2016syntax},  both of these assume syntactic knowledge is provided by external models during training and inference time. }


In this paper, we introduce a novel Transformer encoder- based architecture for NLU tasks with syntactic knowledge encoded that does not require syntactic information to be available during inference time. This is accomplished, first, by training one attention head to predict syntactic ancestors of each token. 
The dependency relationship between each token is obtained from syntactic dependency trees, where each word in a sentence is assigned a syntactic head that is either another word in the sentence or an artificial root symbol \cite{dozat2016deep}. 
\jx{Adding the objective of dependency relationship prediction allows a given token to attend more to its syntactically relevant parent and ancestors.}
\kw{In addition to dependency parsing knowledge, we encode part of speech (POS) information in Transformer encoders because previous research shows that the POS information can help dependency parsing \cite{nguyen2018improved}}.
The closest work to ours is~\cite{strubell2018linguistically}. However, they focused on semantic role labeling and trained one attention head to predict \nr{the} direct parent instead of all ancestors. 

We \kw{compare}
our models with several state-of-the-art neural NLU models on two 
publicly available benchmarking datasets: the ATIS~\cite{hemphill1990atis} and SNIPS~\cite{coucke2018snips} datasets. 
The results show that our models outperform previous
 works.
\kw{To examine the effects of adding syntactic information, we conduct an ablation study and visualize the self-attention weights in the Transformer encoder.}

%% file: problem_definition.tex

We define intent detection (ID) and slot filling (SF) as an utterance-level and token-level multi-class classification task, respectively. 
Given an input utterance with $T$ tokens, we 
predict an intent $y^{int.}$ and a sequence of slots, 
one per token, $\{y^{slot}_{1}, y^{slot}_{2}, \dots, y^{slot}_{T}\}$ as outputs. We add an empty slot denoted by ``O" to represent words containing no labels.
The goal is to maximize the likelihood of correct the intents and slots given input utterances.  



%% file: proposed_model.tex
\begin{figure}[t]
    \centering
    \includegraphics[width=0.48\textwidth]{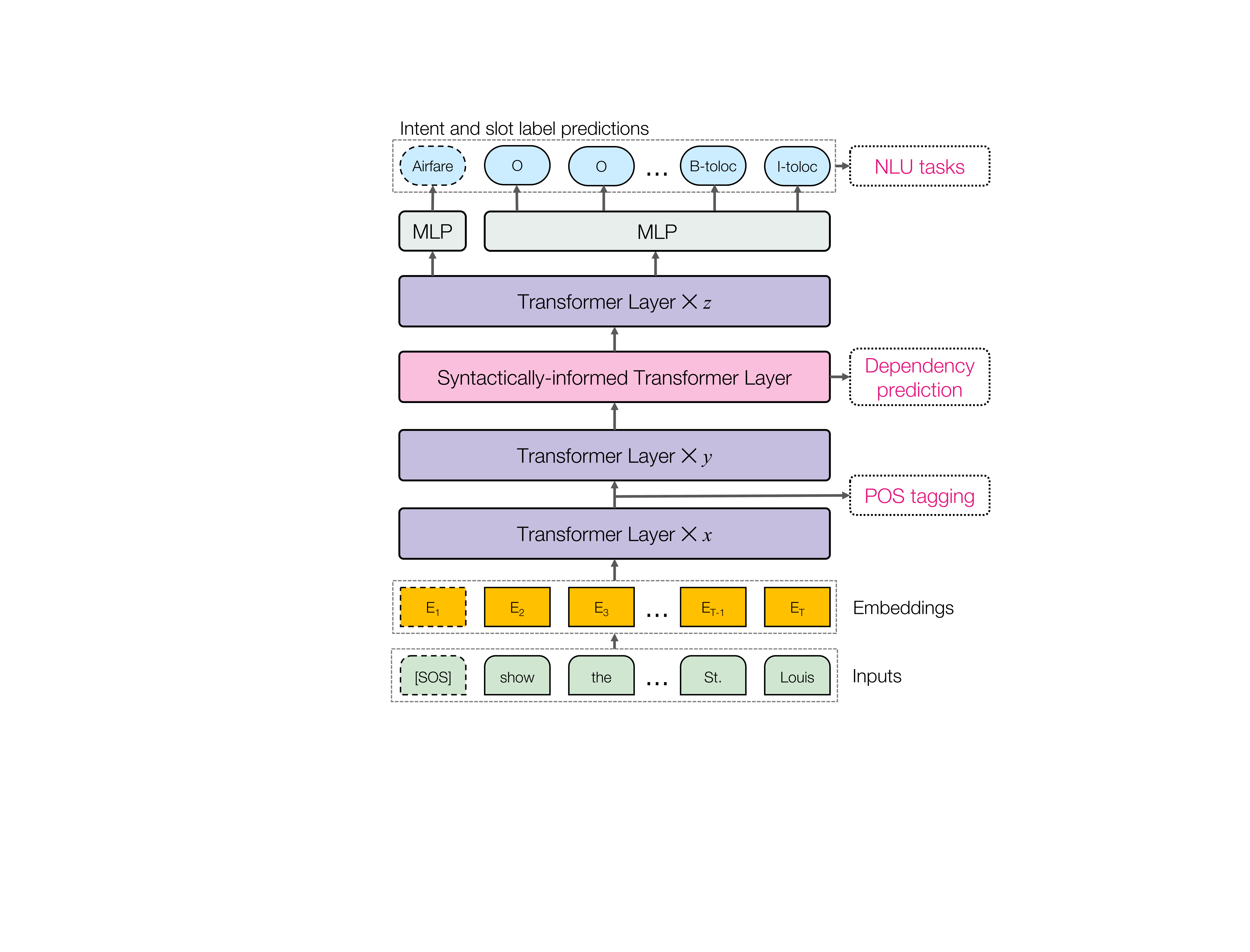} 
    \caption{A high level overview of the proposed architecture.  
    Note that $x$, $y$ and $z$ all refer to number of layers that can vary depending on implementation. ``MLP'' refers to a multi-layer perceptron (MLP).
    }
    \label{overview}
\end{figure}

We jointly train our model for NLU tasks (\ie, ID and SF), syntactic dependency prediction and POS tagging via multi-task learning~\cite{caruana1993multitask}, as shown in Figure~\ref{overview}.
For dependency prediction, we insert a syntactically-informed Transformer encoder layer after the ($x+y$)th layer.  
In this encoder layer, one attention head is trained to predict the full ancestry for each token on the dependency parsing tree.
For POS tagging,
we add a 
POS 
\jx{tagging model}
\jx{that shares the first $x$ Transformer encoder layers with the NLU} model. We describe the details of our proposed architecture below. 

\subsection{Input Embedding}
The input embedding model maps a sequence of token representations $\{t_1, t_2, \dots, t_T\}$ into a sequence of continuous embeddings $\{e_0, e_1, e_2, \dots, e_T\}$, with $e_0$ being the embedding of a special start-of-sentence token, ``[SOS]''. 
The embeddings are then fed into the NLU model.

\subsection{Transformer Encoder Layer}
\jx{The Transformer encoder layers are originally proposed in~\cite{vaswani2017attention}.}
Each encoder layer consists of a multi-head self-attention layer and feed forward layers with layer normalization and residual connections. We stack multiple encoder layers to learn contextual embeddings of each token, each with $H$ attention heads. Suppose the output embeddings of the encoder layer $j-1$ is $E^{(j-1)}$, each attention head $h$ at layer $j$ first calculates self-attention weights by the scaled dot product (\ref{eq:1}).
\begin{equation}
    A_h^{(j)} = softmax(\frac{Q^{(j)}_h{K^{(j)}_h}^T}{\sqrt{d_k}})
    \label{eq:1}
\end{equation}
In (\ref{eq:1}), the query $Q^{(j)}_h$ and key $K^{(j)}_h$ are two different linear transformations of $E^{(j-1)}$, $d_k$ is the dimension of the query and the key embeddings.
The output of the attention head $h$ is calculated by:
\begin{equation}
    F_h^{(j)} = A_h^{(j)} V_h^{(j)}
\end{equation}
\jx{in which the value $V_h^{(j)}$ is also a linear transformation of $E^{(j-1)}$.}
The outputs of $H$ attention heads are concatenated as the self-attended token representations, followed by another linear transformation:
\begin{equation}
    Multi^{(j)} = [F_1^{(j)}, F_2^{(j)}, \cdots, F_H^{(j)}]W^F
\end{equation}
which is fed into the next feed forward layer. Residual connections and layer normalization are applied after the multi-head attention and feed forward layer, respectively. 


\begin{figure*}[t]
    \centering
    \includegraphics[width=0.9\textwidth]{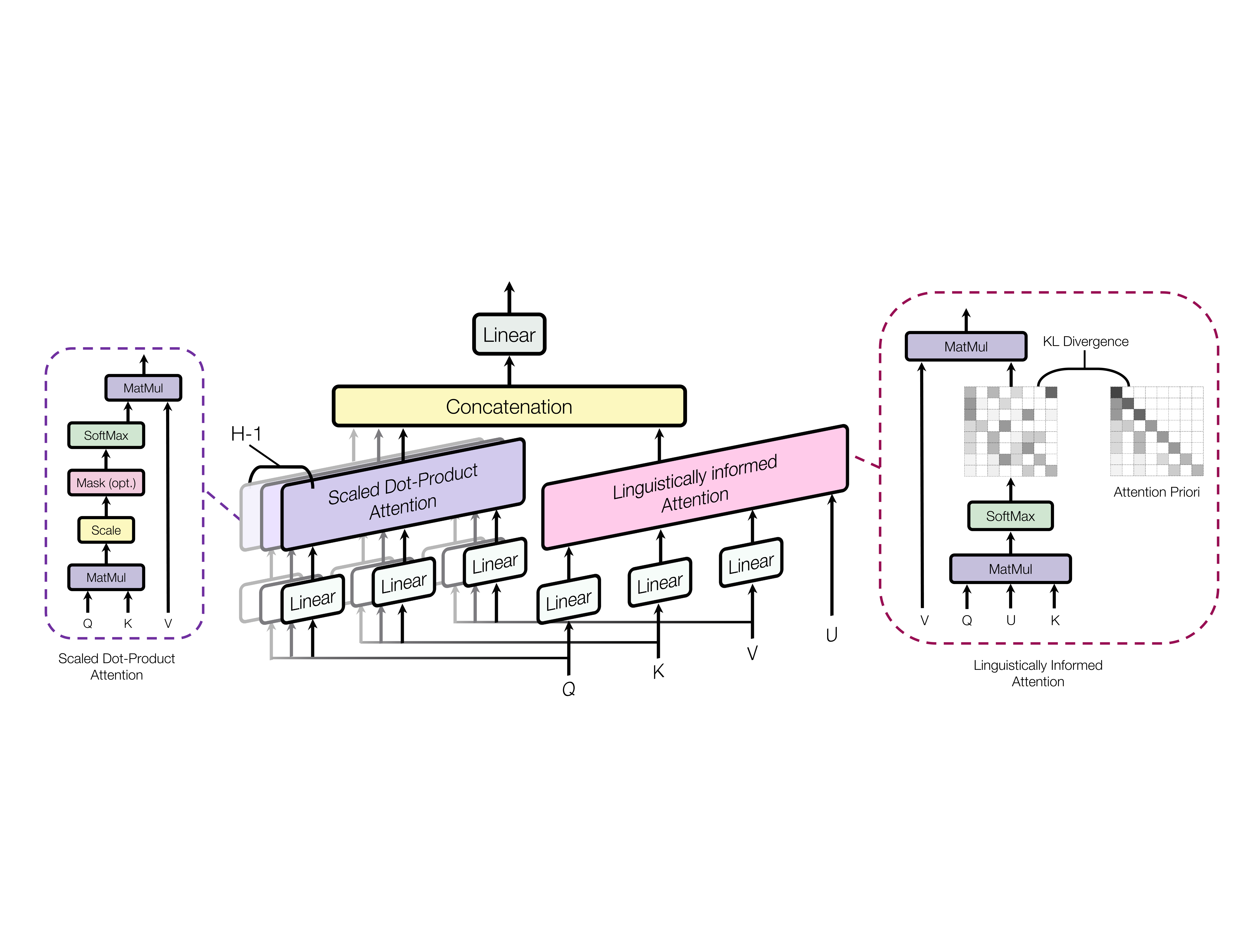} 
    \caption{Overview of the syntactically-informed Transformer layer. One out of the $H$ attention heads is trained for predicting syntactic parse ancestors of each token. For each token, this attention head outputs a distribution over all positions in the sentence, which corresponds to the probability of each token being the ancestor of this token.
    The loss function is defined as the mean \nr{Kullback–Leibler (KL)} divergence between the output distributions of all tokens and the corresponding prior distributions. }
    \label{transformer}
\end{figure*}

\subsection{Encoding Syntactic Dependency Knowledge}
\label{dep}

As shown in Figure~\ref{transformer}, the syntactically-informed transformer encoder layer differs from the \nr{standard}
Transformer encoder layer by having one of the $H$ attention heads trained
to predict the full ancestry for each token, \ie, parents, grandparents, great grandparents, \etc.
\jx{Different from ~\cite{strubell2018linguistically}, we use full ancestry prediction instead of just direct parent prediction.}
Later we will demonstrate the benefits of our approach
\jx{in the Results section}.  

Given an input sequence of length $T$, the output of a regular attention head is a $T \times T$ matrix, in which each row contains the attention weights that a token puts on all the tokens in the input sequence. The output of the syntactically-informed attention head is also a $T \times T$ matrix but this attention head is trained to assign weights only on the syntactic governors (\ie, ancestors) of each token. 

To train this attention head, 
we define a loss function by the difference between the output attention weight matrix of 
the syntactically-informed attention head and a predefined prior attention weight matrix. The prior attention weight
\jx{matrix} contains the prior knowledge that each token should attend to its syntactic parse ancestors, with attention weight\jx{s} being higher on ancestors that are closer to that token.
During training, we obtain the prior attention weights based on the outputs of a pre-trained dependency parser.

\begin{figure}[h]
    \centering
    \includegraphics[width=0.4\textwidth]{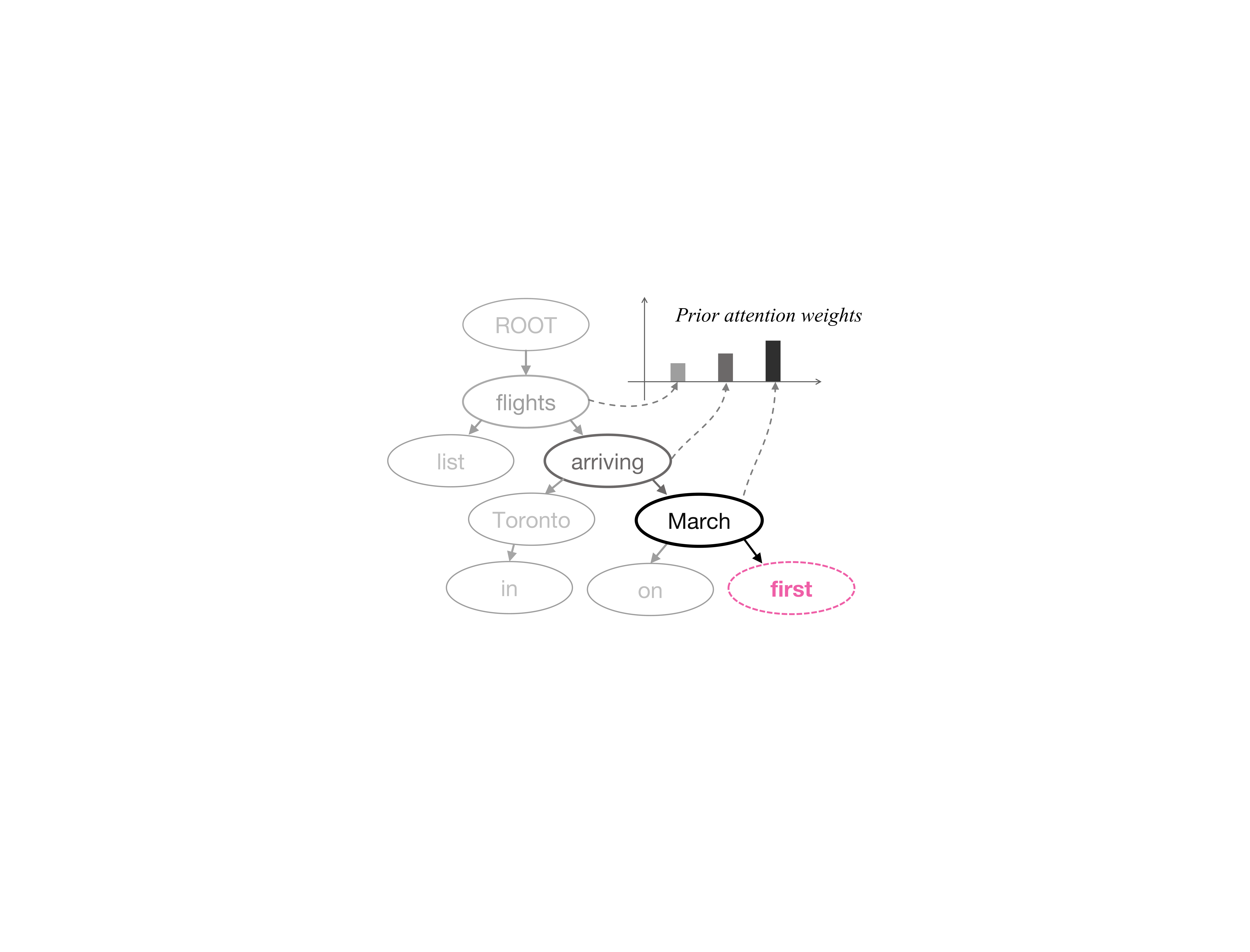} 
    \caption{Syntactic dependency tree of ``\textit{list flights arriving in Toronto on March first}'' and prior attention weights of word ``\textit{first}''. }
    \label{prior}
\end{figure}

For example, in the utterance ``\textit{list flights arriving in Toronto on March first}'', the syntactic parse ancestors of word ``\textit{first}'' are ``\textit{March}'', ``\textit{arriving}'' and ``\textit{flights}'' which are $1$, $2$ and $3$ hops away on the dependency tree, respectively, as shown in Figure~\ref{prior}. The ancestors are syntactically meaningful for the determination of the slot of ``\textit{first}'', which is ``\textit{arrive date, day number}'' in this case.

To train the attention head to assign higher weight\jx{s} on the ancestors of each token, we define 
prior attention weight\jx{s} of each token based on the distance between the token and its ancestors. Formally, the prior attention weight\jx{s} of token $i$ \jx{are} defined as:
\begin{equation}
    w_{i, j}^{prior}=\left\{
                \begin{array}{ll}
                  softmax(- d_{i, j} / \tau) & \text{if $j \in ancestors(i)$} \\
                  0 & \text{if $j \notin ancestors(i)$}
                \end{array}
              \right.
\end{equation}
\jx{in which $d_{i,j}$ is the distance between token $i$ and $j$, $softmax$ is the Softmax function, $\tau$ is the temperature of the Softmax function controlling the variance of the attention weights over all the ancestors and $i, j \in \{1, 2, \dots, T\}$.}
The stack of prior attention weight\jx{s} of all $T$ tokens is a $T \times T$ matrix, denoted by $W^{prior}$.
We train our model to decrease the difference between $W^{prior}$ and the attention matrix $W^{(s)}_h$ output by the attention head $h$ at the $s$th layer.
The difference is measured by the mean of row-wise KL divergence between these two matrices, which is used as \jx{an additional} loss function \jx{besides the NLU loss functions}. We refer to this loss as dependency loss denoted by $\mathcal{L}^{dep.}$,
formally:

\begin{align}
    \mathcal{L}^{dep.} & = \frac{1}{T}\sum_{i=1}^T \infdiv{W^{prior}_i}{W^{(s)}_{h, i}} \\
    W^{(s)}_h & = softmax ( Q^{(s)} U^{dep.} (K^{(s)})^T )
    \label{eq:6}
\end{align}
in which $D_{KL}(\cdot{})$ denotes the KL divergence, $Q^{(s)}$ and $K^{(s)}$ are linear transformations of $E^{(s-1)}$, $W^{(s)}_{h,i}$ is the $i$th row of $W^{(s)}_h$, and $U^{dep.}$ is a parameter matrix.
\jx{
\kw{I}n (\ref{eq:6}) we use the biaffine attention instead of the scaled dot product attention, which has been shown to be effective for dependency parsing~\cite{dozat2016deep}.
}

We treat $\tau$ as a hyperparameter and tune it on \jm{the} validation set.
With $\tau \to 0^{+}$, attention head $h$ will be trained to only pay attention to the direct parent of each token, a special case used by~\cite{strubell2018linguistically}.
Thus, our method is a more general approach compared to~\cite{strubell2018linguistically}.

\subsection{Encoding Part-of-Speech Knowledge}
\label{pos}
Part-of-Speech (POS) information is important for disambiguating words with multiple meanings \cite{alva2016hidden}. This is because an ambiguous word carries a specific POS in a particular context \cite{pal2015approach}.
For instance, the word ``\textit{May}'' could be either a verb or a noun. Being aware of its POS tag is beneficial for down-stream tasks, such as predicting the slots in the utterance ``\textit{book a flight on May 1st}''.
Furthermore, previous studies have shown that while models trained for a sufficiently large number of steps can potentially learn underlying patterns of POS, the knowledge is imperfect \cite{jawahar2019does, sundararaman2019syntax}. 
For these reasons, we explicitly train our model to perform POS tagging using the POS tags generated by 
\jx{a pretrained POS tagger.}

Similar to slot filling, we simplify POS tagging as a token-level classification problem. 
We apply a MLP-based classifier on the output embeddings of the $r$th transformer encoder layer and use cross entropy as the loss function:

\begin{align}
    p_{i, o}^{pos} &= softmax(MLP(e_{r, i})) \\
    \mathcal{L}^{pos} &= -\sum_{i=1}^T \sum_{o=1}^O y_{i, o}^{pos} \log p_{i, o}^{pos}
\end{align}
in which $p_{i,o}^{pos}$ is the predicted probability of the $i$th token's POS label being the $o$th label in the POS label space, $O$ is the total number of POS labels, $y_{i,o}^{pos}$ is the one-hot representation of the groundtruth POS label.

\subsection{Intent Detection and Slot Filling}
\subsubsection{Intent detection:} We apply a linear classifier on the embedding of the ``[SOS]'' token, $e_{L, 0}$, which is output by the last Transformer encoder layer $L$. Cross entropy loss is used for intent detection. The loss on one utterance is defined as:

\begin{align}
    p_{n}^{int.} &= softmax(W^{int.} e_{L, 0}^T + b^{int.}) \\
    \mathcal{L}^{int.} &= - \sum_{n=1}^N y_{n}^{int.} \log p_{n}^{int.}
\end{align}
in which $W^{int.}$ and $b^{int.}$ are the parameters of the linear classifier, $y^{int.}$ is the one-hot representation of the ground truth intent label, $N$ is the total number of intent labels and $p_n^{int.}$ is the predicted probability of this utterance's intent label being the $n$th label in the intent label space.

\subsubsection{Slot filling:}
\jm{We apply a MLP-based classifier on the embeddings output by the last Transformer encoder layer using cross entropy as the loss function.}
The loss on one utterance is defined as follow:

\begin{align}
    p_{i, s}^{slot} &= softmax(MLP(e_{L, i})) \label{eq:11} \\
    \mathcal{L}^{slot} &= -\sum_{i=1}^T \sum_{s=1}^S y_{i, s}^{slot} \log p_{i, s}^{slot}
\end{align}
in which $p_{i,s}^{slot}$ is the predicted probability of the $i$th token's slot being the $s$th label in the slot space, $S$ is the total number of slots, and $y_{i,s}^{slot}$ is the one-hot representations of the ground truth slot label.

\subsection{Multi-task Learning}
We train our model via multi-task learning~\cite{caruana1993multitask}. Our loss function is defined as:
\begin{equation}
    \mathcal{L} = \mathcal{L}^{NLU} + c^{dep} \cdot \mathcal{L}^{dep} + c^{pos} \cdot \mathcal{L}^{pos}
\end{equation}
\kw{where} 
$\mathcal{L}^{NLU}$ equals to $\mathcal{L}^{slot}$ for slot filling, $\mathcal{L}^{int.}$ for intent detection or $\mathcal{L}^{slot} + \mathcal{L}^{int.}$ for joint training, and $c^{dep}$ and $c^{pos}$ are coefficients of 
the dependency prediction loss and the POS tagging loss, respectively.
$c^{dep}$ and $c^{pos}$ are treated as hyperparameters and selected based on validation performance.

\begin{table*}[t]
    \centering
    \begin{tabular}{ccccccc}
      & \multicolumn{2}{c}{SNIPS} & & \multicolumn{3}{c}{ATIS} \\ 
     \cline{2-3} \cline{5-7}
       & SF & ID & & SF & ID-M & ID-S  \\ \hline\hline
    Joint Seq~\cite{hakkani2016multi}           & 87.30          & 96.90          & & 94.30          & 92.60          &   -            \\ 
    Attention-based RNN~\cite{liu2016attention} & 87.80          & 96.70          & & 95.78          &   -            & 97.98          \\ 
    Slot-Gated ~\cite{goo2018slot}              & 89.27          & 96.86          & & 95.42          & 95.41          &   -            \\ 
    SF-ID, SF first~\cite{e2019a}               & 91.43          & 97.43          & & 95.75          & 97.76          &   -            \\ 
    SF-ID, ID first~\cite{e2019a}               & 92.23          & 97.29          & & 95.80          & 97.09          &   -            \\ 
    Stack-Propagation~\cite{qin2019a}           & 94.20          & 98.00          & & 95.90          & 96.90          &   -            \\ 
    Graph LSTM~\cite{zhang2020graph}            & 95.30          & 98.29          & & 95.91          & 97.20          &   -            \\ 
    TF                                          & 96.37          & 98.29          & & 95.31          & 96.42          & 97.65          \\ 
    \hline                                                                                                              
    \textbf{SyntacticTF (Independent)}          & 96.56          & 98.71          & & 95.94          & \textbf{97.76} & 98.10          \\ 
    \textbf{SyntacticTF (Joint)}        & \textbf{96.89} & \textbf{99.14} & & \textbf{96.01} & 97.31          & \textbf{98.32} \\ 
       
    \hline
   JointBERT~\cite{chen2019bert}$^*$      & 97.00 & 98.60  & & 96.10 & 97.50 &  -   \\ 
    \end{tabular}
    \caption{SF and ID results on the ATIS and SNIPS dataset (\%). \textbf{TF} refers to the Transformer encoder-based model trained without syntactic information. \textbf{SyntacticTF} refers to our model.
    \textbf{Independent} and \textbf{Joint} refer to independently and jointly training for SF and ID, respectively. \textbf{ID-M} refers to multiple label matching for intent detection evaluation and \textbf{ID-S} refers to single label matching.
    $^*$This work relies on pretraining, which is not required by other works in the table.}

    \label{nlu_results}
\end{table*}

%% file: experiment.tex
\subsection{Datasets}
We conducted experiments on two 
benchmark datasets: the Airline Travel Information Systems (ATIS)~\cite{hemphill1990atis} and SNIPS~\cite{coucke2018snips} datasets.
The ATIS dataset has a focus on airline information and has been used as benchmark on NLU tasks.
We used the same version as~\cite{goo2018slot,e2019a} that contains 4,478 utterances for training, 500 for validation and 893 for testing.
The SNIPS dataset has a focus on personal assistant commands, with a larger vocabulary size and more diverse intents and slots.
It contains 13,084 utterances for training, 700 for validation and 700 for testing.

\subsection{Evaluation Metrics}
We use classification accuracy for intent detection and the F1 score for slot filling, which is the harmonic \kw{mean}
of precision and recall. 
For the SNIPS dataset, we use the same version and evaluation method as pervious works~\cite{zhang2020graph}.
For the ATIS dataset, we \kw{find}
that previous works \kw{use}
two different evaluation methods for intent detection on utterances with multiple labels.
The first method 
count\kw{s} a prediction as correct if it is equal to one of the ground truth labels of the utterance
 ~\cite{liu2016attention}. We refer this method as the single label matching method (ID-S).
The second method \kw{counts}
a prediction as correct only if it matches all labels of the utterance
~\cite{goo2018slot,e2019a}. We refer this method as the multiple label matching method (ID-M).
\kw{We report both in our results.}

\subsection{Implementation Details}
Our experiments are implemented in PyTorch~\cite{paszke2017automatic}. 
The hyperparameters are selected based on the performance on the validation set. We use the Adam optimizer~\cite{kingma2015adam} with $\beta_1=0.9$, $\beta_2=0.999$, $\epsilon = 10^{-7}$ and the weight decay fix as described in~\cite{loshchilov2017decoupled}. Our learning rate schedule first increases the learning rate linearly from 0 to 0.0005 (warming up) and then decreases it to 0 following the values of the cosine function.
We use warming up steps $\approx 20\%$ of the total training steps. The specific number of warming up steps is determined by validation performance. We use the implementation of the optimizer and learning rate scheduler of the Transformers library~\cite{Wolf2019HuggingFacesTS}.

We use Stanza~\cite{qi2020stanza} to generate training labels for POS tagging and dependency prediction.
For the NLU model trained with both dependency prediction and POS tagging, $c^{dep}$ and $c^{pos}$ are both set to $1$. 
For the NLU model trained with only dependency prediction, $c^{dep}$ is set to $5$.
We used weight decay of 0.1 and dropout rate~\cite{srivastava2014dropout} of 0.1 and 0.3 for the SNIPS and ATIS dataset, respectively. We use batch size of 32 and train each model for $100$ epochs. We report the testing results of the checkpoints achieving the best validation performance.

We use the concatenation of GloVe embeddings~\cite{pennington2014glove} and character embeddings~\cite{hashimoto2017joint} as token embeddings and keep them frozen during training.
The hidden dimension of the Transformer encoder layer is 768 and the size of feed forward layer is 3072. Considering the small size of the two datasets, we only use two Transformer encoder layers in total (with $x=1$, $y=0$ and $z=0$ as in Figure~\ref{overview}), each of which has $4$ attention heads.
The dimension of $Q^{(s)}$ and $K^{(s)}$ is $200$.
For slot filing, we apply the Viterbi decoding at test time.
BIO is a standard format for slot filling annotation schema, as shown in Figure~\ref{fig:example}.
The transition probabilities are manually set to ensure the output sequences of BIO labels to be valid, by simply specifying the probabilities of invalid transition to zero and the probabilities of valid transition to one.

\subsection{Baseline Models}
We compare our proposed model with the following baseline models:
\begin{itemize}
    \item Joint Seq~\cite{hakkani2016multi} is a joint model for intent detection and slot filling based on the bi-directional LSTM model.         
    \item Attention-based RNN~\cite{liu2016attention} is a sequence-to-sequence model with the attention mechanism.
    \item Slot-Gated ~\cite{goo2018slot} utilizes intent information for slot filling through the gating mechanism.            
    \item SF-ID \cite{e2019a} is an architecture that enables the interaction between intent detection and slot filling.
    \item Stack-Propagation~\cite{qin2019a} is a joint model based on the Stack-Propagation framework.
    \item Graph LSTM~\cite{zhang2020graph} is based on the Graph LSTM model. 
    \item JointBERT~\cite{chen2019bert} is a joint NLU model fined tuned from the pretrained BERT model~\cite{devlin2018bert}. 
    \item TF is the Transformer encoder-based model trained without syntactic information.
\end{itemize}

%% file: results.tex
\kw{Table~\ref{nlu_results} shows the performance of 
\jx{the} baseline and proposed models for SF and ID on the SNIPS and ATIS
\jx{dataset}, respectively. Overall, our proposed models achieve the best performance on 
\jx{the two}
benchmarking datasets. On the SNIPS dataset, our proposed joint model achieves an absolute \jx{F1 score and}
accuracy improvement of $1.59\%$ and $0.85\%$ for SF and ID, respectively, compared to the best performed baseline model without pre-training
\jx{~\cite{zhang2020graph}}.
On the ATIS dataset, our proposed joint model also achieves an absolute F1 score
\jx{and accuracy}
improvement of $0.1\%$ and $0.34\%$ for SF and ID-S, compared to the best performed baseline model for SF\jx{~\cite{zhang2020graph}}
and 
ID-S\jx{~\cite{liu2016attention}}, respectively.
In addition, our proposed independent model achieves the same performance as the best performed baseline model on ID-M\jx{~\cite[SF-ID, SF first]{e2019a}}.
}

\jx{Besides, the model based on Transformer encoder without syntactic knowledge can achieve SOTA results on the SNIPS dataset and is slightly worse than the SOTA results on the ATIS dataset. 
This indicates the powerfulness of the Transformer encoder for SF and ID.
Moreover, the further improvement of our models over the baseline models demonstrates the benefits of incorporating syntactic knowledge. Additionally, compared to previous works with heterogenous model structures, our models are purely based on self-attention and feed forward layers. }

\kw{We also find that our proposed models can outperform
\jx{the JointBERT} model with pre-training~\cite{chen2019bert} for intent detection tasks. Compared to the
\jx{JointBERT} model, our proposed joint model achieves an absolute accuracy improvement of $0.54\%$ for ID on the SNIPS dataset; and our proposed \jx{independent} model achieves an absolute accuracy improvement of $0.26\%$ for ID-M on the ATIS dataset. While our proposed model does not outperform the
JointBERT model for SF, the performance gap is relatively small ($~0.11\%$ on SNIPS and $~0.09\%$ on ATIS). It should be noted that
\jx{our model does not require pre-training and} 
the size of our model is only one \jx{seventh} of the JointBERT model (16 million vs. 110 million parameters).}

\jx{Previous works have shown that models like BERT can learn syntactic knowledge by self-supervision~\cite{clark2019does, manning2020emergent}. This can partially explain why the JointBERT can achieve very good results without being fed with syntactic knowledge explicitly.
}

\subsection{Ablation Study}
\kw{Table~\ref{tbl:ablation} shows the ablation study results of the effects of adding different syntactic information.}
\kw{A first observation is that t}he model trained with a singe syntactic task, either dependency prediction or POS tagging, outperforms the baseline Transformer encoder-based model without syntactic information.
\kw{This gives us confidence that syntactic information can help improve model performance.}
Moreover, training a Transformer model with both the syntactic tasks achieves even better results than training with a single syntactic task. \kw{This could be because the POS tagging task improves the performance of the dependency prediction task \cite{nguyen2018improved}, which in turn improves the performance of SF and ID. }

\begin{table}[t]
    \centering
    \begin{tabular}{ccccccc}
      & \multicolumn{2}{c}{SNIPS} & & \multicolumn{3}{c}{ATIS} \\ 
     \cline{2-3} \cline{5-7}
       & SF & ID & & SF & ID-M & ID-S  \\ \hline\hline
    \textbf{TF}                     & 96.37          & 98.29          & & 95.31          & 96.42          & 97.65            \\ 
    \textbf{TF + D}                 & 96.31          & 98.43          & & \textbf{95.99} & 96.53          &  \textbf{98.76}  \\ 
    \textbf{TF + P}                 & 96.47          & 98.57          & & 95.82          & 97.31          & 98.10            \\ 
   \textbf{TF + D + P}              & \textbf{96.56} & \textbf{98.71} & & 95.94          & \textbf{97.76} & 98.10            \\ 
    \end{tabular}
    \caption{Results of ablation study. \textbf{TF} refers to the baseline models with two Transformer encoder layers. \textbf{D} and \textbf{P} refers to dependency prediction and POS tagging, respectively.}
    \label{tbl:ablation}
\end{table}


Interestingly, we observe that the addition of dependency prediction reduces the performance of slot filling on the SNIPS dataset ($96.31\%$) when compared to the baseline Transformer encoder-based model ($96.37\%$).  There are several potential reasons.
Firstly, the sentences in the SNIPS dataset are overall shorter than the ATIS dataset so that
\jx{the syntactic dependency information}
might be less helpful.
\jx{
Secondly, previous work has shown that syntactic parsing performance often suffers when a named entity span has crossing brackets with the spans on the parse tree \cite{finkel2009joint}.
Thus, the dependency prediction performance of our model might decrease due to the presence of many name entities in the SNIPS dataset, such as song names and movie names, which could introduce noisy dependency information into the attention weights and degrade the performance on the NLU tasks.
}
\subsection{Qualitative Analysis}
We qualitatively examined the
errors made by \nr{the} Transformer encoder-based models with and without syntactic information 
to understand in what ways syntactic information helps improve the performance.
Our major findings are:

\textbf{ID errors related to preposition with nouns:}
Prepositions, when appearing between nouns, are used to describe their relationship. 
For example, in the utterance ``\textit{kansas city to atlanta monday morning flights}", the preposition ``\textit{to}''
\jx{denotes} the direction from ``\textit{kansas city}" (departure location, noun) to ``\textit{atlanta}'' (arrival location, noun). 
Without this knowledge, 
\jx{a}
model could misclassify the intent of this utterance as asking for city information 
rather than flight information. We found that about $50\%$ of the
\jx{errors made by the model without syntactic information}
contain this pattern,
whereas less than $10\%$ of the misclassified utterances contain this pattern for the model with syntactic information (See Appendix A for the full list).

\textbf{SF errors due to POS confusion:}
A Word can have multiple meanings depending on context. 
For example, the same word ``\textit{may}" can be a verb expressing possibility, or as a noun referring to the fifth month of the year. 
We found that correctly recognizing the POS of words could potentially help reduce slot filling errors. For example, in this utterance ``\textit{May I have the movie schedules for Speakeasy Theaters}", the slot for ``\textit{May}" should be empty, but the model without syntactic information predicts it as ``\textit{Time Range}''. 
By contrast, the model with syntactic information predicts correctly for this word,
probably because the confusion of noun vs. verb for the word ``\textit{May}'' is addressed by incorporating POS information. More examples are included in Appendix A.



\begin{table}[t]
    \centering
    \begin{tabular}{cccccc}
      & \multicolumn{2}{c}{ATIS} & & \multicolumn{2}{c}{SNIPS} \\ 
     \cline{2-3} \cline{5-6}
       & SF & ID & & SF & ID-S  \\ \hline\hline
    \textbf{TF + Par.}                 & 96.20 &  98.29 & & 95.58 & 98.10   \\
    \textbf{TF + Anc.}                 & 96.31 &  98.43 & & 95.99 & 98.76   \\

    \end{tabular}
    \caption{Intent detection and slot filling results of Transformer (\textbf{TF}) encoder-based models with dependency parent prediction (\textbf{Par.}) and dependency ancestor prediction (\textbf{Anc.}) on the ATIS and SNIPS dataset.}
    \label{tbl:dep_results}
\end{table}
\subsection{Parent Prediction vs. Ancestor Prediction}
\jx{
We compare our approach of predicting all ancestors of each token with the approach described in~\cite{strubell2018linguistically}, which only predicts direct dependency parent of each token.
Results in Table~\ref{tbl:dep_results} show that the model with our approach can achieve better results for both ID and SF on the two datasets, which demonstrates that our approach is more beneficial to the NLU tasks.
We hypothesize that incorporating syntactic ancestor prediction can better capture long-distance syntactic relationship.
As shown in~\cite{tur2010what}, long distance dependencies are important for slot filling.   
For example, in the utterance ``\textit{Find flights to LA arriving in no later than next Monday}'', a 6-gram context is needed to figure out that ``\textit{Monday}'' is the arrival date instead of the departure date.
}

\begin{figure}[t]
    \centering
    \includegraphics[width=0.4\textwidth]{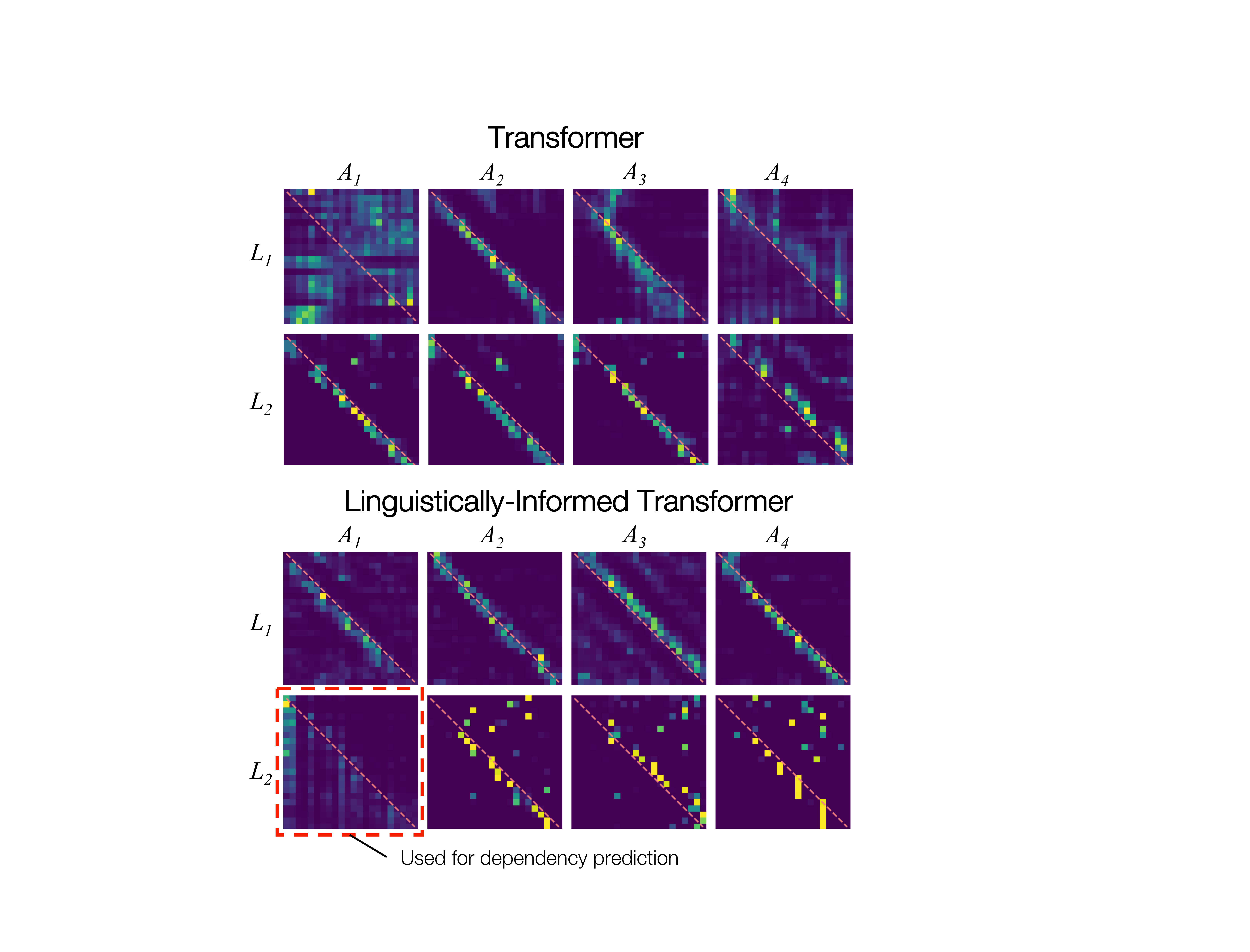} 
    \caption{Visualization of the attention weights of \nr{the} model with and without syntactic supervision for slot filling. $L_i$ and $A_j$ stands for the $i$th Transformer layer and $j$th attention head, respectively. The attention head inside the \nr{red-dotted} box is trained for dependency prediction.  }
    \label{fig:attention}
\end{figure}

\subsection{Visualization of Attention Weights}
We visualize the attention weights output by models trained with and without syntactic information to understand what the models have learned by incorporating syntactic information. 
We select the utterance ``\textit{show me the flights on american airlines which go from st. petersburg to ontario california by way of st. louis}" from the ATIS testing set. 
Only the model trained with syntactic information predicts the slot labels correctly. 
As shown in Figure~\ref{fig:attention}, the model without syntactic information \nr{has} simple attention patterns on both layers, such as looking \nr{backward} and looking forward. Other attention heads seem to be random and less informative.

In contrast, the model with syntactic information has more informative attention patterns. On the first layer, all the attention heads present simple but diverse patterns. Besides looking forward and backwards, the second attention head looks at both directions for each token. On the second layer, however, we observe more complex patterns and long-distance attention which could account for more task-oriented operations. Therefore, it is possible that the Transformer encoder learns attention weights 
\nr{better}
with syntactic information supervision so that the encoder can leave more power for the end task.

%% file: related_work.tex
Research on intent detection and slot filling 
emerged in the $1990s$ from the call classification systems \cite{gorin1997may} and the ATIS project \cite{price1990evaluation}. Early work has primarily focused on using \nr{a} traditional machine learning classifier such as CRFs 
\cite{haffner2003optimizing}. Recently, there has been an increasing application of neural models on NLU tasks. These approaches, primarily based on RNNs, have shown that neural approaches outperform traditional models \cite{ mesnil2014using, tur2012towards, zhang2016joint, goo2018slot, e2019a}. For example, Mesnil et al (2015) employed RNNs for slot filling and found an $2.3\%$ relative improvement of F1 compared to CRF \cite{mesnil2014using}. 
\jm{Some works} also explored Transformer encoder and graph LSTM-based neural architectures \cite{chen2019bert, zhang2020graph}.

%
Syntactic information has been shown to be beneficial to many tasks, such as neural machine translation~\cite{akoury2019syntactically}, semantic role labeling~\cite{strubell2018linguistically}, and machine reading comprehension~\cite{zhang2020sg}.
Research on NLU tasks has also shown that incorporating syntactic information into machine learning models can help improve the performance.
Moschitti \etal~\shortcite{moschitti2007spoken} used syntactic information for slot filling, where the authors used a tree kernel function to encode the structural information acquired by a syntactic parser.
An extensive analysis on the ATIS dataset revealed that most NLU errors are caused by complex syntactic characteristics, such as prepositional phrases and long distance dependencies~\cite{tur2010what}. Tur \etal~\shortcite{tur2011sentence} proposed a rule-based dependency parsing based sentence simplification method to \nr{augment}
the input utterances based on the syntactic structure.
Compared to previous works, \kw{our work is the first to encode syntactical knowledge into end-to-end neural models for intent detection and slot filling.}

%% file: conclusion.tex
In this paper, we propose to encode syntactic knowledge into the Transformer encoder-based model for intent detection and slot filling.
Experimental results indicate that a model with only two Transformer encoder layers can already match or even outperform the SOTA performance on two benchmark datasets.
Moreover, we show that the performance of this baseline model can be further improved by incorporating syntactical supervision.
The visualization of the attention weights also reveals that syntactical supervision can help the model to better learn syntactically-related patterns.
For future work, we will evaluate our approach with larger model sizes on larger scale datasets containing more syntactically complex utterances.
Furthermore, we will investigate incorporating syntactic knowledge into models pretrained by self-supervision and applying those models on the NLU tasks.